\pdfoutput=1
\documentclass[11pt]{article}
\usepackage[preprint]{acl}
\usepackage{times}
\usepackage{latexsym}
\usepackage[T1]{fontenc}
\usepackage[utf8]{inputenc}
\usepackage{microtype}
\usepackage{inconsolata}
\usepackage{booktabs}
\usepackage{multirow}
\usepackage{xcolor}
\usepackage{graphicx}
\usepackage{amsmath}

\newcommand{\bench}{\textsc{Puzzling-Bench}}
\newcommand{\pilot}{\textsc{Pilot-255}}
\newcommand{\imgpilot}{\textsc{Img-100}}

\title{\bench{}: Evaluating Language Models on 23,777 Human-Written Puzzles with Community-Verified Answers}

\author{
  \textbf{Abhinav Sharma\textsuperscript{1}},
  \textbf{Li Li\textsuperscript{2}},
  \textbf{Chien Van Nguyen\textsuperscript{3}},
  \textbf{Sai Karthik Navuluru\textsuperscript{4}},
\\
  \textbf{Bo Ni\textsuperscript{5}},
  \textbf{Xiangbo Gao\textsuperscript{6}},
  \textbf{Xiaolei Huang\textsuperscript{7}},
  \textbf{Yue Zhao\textsuperscript{2}},
  \textbf{Nesreen K. Ahmed\textsuperscript{8}},
\\
  \textbf{Tyler Derr\textsuperscript{5}},
  \textbf{Yu Wang\textsuperscript{9}},
  \textbf{Hongjie Chen\textsuperscript{10}},
  \textbf{Zhengzhong Tu\textsuperscript{6}},
  \textbf{Thien Huu Nguyen\textsuperscript{3}},
\\
  \textbf{Ryan A. Rossi\textsuperscript{11}},
  \textbf{Franck Dernoncourt\textsuperscript{12}}
\\
\\
  \textsuperscript{1}University of Massachusetts Amherst,
  \textsuperscript{2}University of Southern California,
\\
  \textsuperscript{3}University of Oregon,
  \textsuperscript{4}University of Texas at Dallas,
  \textsuperscript{5}Vanderbilt University,
\\
  \textsuperscript{6}Texas A\&M University,
  \textsuperscript{7}University of Missouri,
  \textsuperscript{8}Cisco,
  \textsuperscript{9}University of Georgia,
\\
  \textsuperscript{10}Dolby Laboratories,
  \textsuperscript{11}Zillow,
  \textsuperscript{12}Adobe Research
\\
  \small{\href{mailto:abhinavs@umass.edu}{\texttt{abhinavs@umass.edu}}}
}

\begin{document}
\maketitle

\begin{abstract}
We introduce \bench{}, a benchmark of 23,777 puzzles with community-verified answers drawn from the Puzzling Stack Exchange archive (2014--2026): riddles, ciphers, logic grids, lateral-thinking problems, wordplay and recreational mathematics, a third of them with images. Every puzzle carries the answer accepted by its author, a short reference answer taken from it (LLM-extracted for the evaluation sets and checked on a 50-puzzle human review), the author's own hints, community tags and a creation date, so that the benchmark supports hint ablations, category analysis and a rolling post-cutoff slice (254 puzzles posted after the training cutoff of six of the nine models we evaluate, including the best one) as a contamination check. We evaluate nine frontier and open models with a two-tier scorer (normalized exact match, then an LLM judge) and validate the judge with three judges from two model families: they agree on 95--97\% of verdicts ($\kappa$ 0.82--0.96), move no model by more than 1.6 points, and show no own-family preference. The best model, GPT-6.1, reaches 74.8\% on a shared 255-puzzle set; a hard core of 17 puzzles defeats all nine models; model failures are strongly correlated, and the second-best model contributes a single puzzle to a greedy ensemble, yet an oracle over all nine reaches 91\%. On a 100-puzzle multimodal pilot, withholding the question's image costs the two best models 24 and 35 points. We release the dataset, the scoring pipeline and all per-row provenance.\footnote{\raggedright Code: \href{https://github.com/a1bhinav/puzzling-bench}{\nolinkurl{github.com/a1bhinav/puzzling-bench}}.\\* Data: \href{https://huggingface.co/datasets/puzzling-bench/puzzling-bench}{\nolinkurl{huggingface.co/datasets/puzzling-bench/puzzling-bench}}.}
\end{abstract}

\section{Introduction}

Puzzles are a natural stress test for language models: each one is short, self-contained, has a single human-verified answer, and requires a combination of world knowledge, lateral thinking and multi-step reasoning that is hard to obtain by pattern matching alone. Puzzling Stack Exchange has accumulated twelve years of such problems. On this site a question is almost always a challenge rather than a request for help: the asker already knows the answer, the community competes to find it, and the asker marks the answer that is correct. The result is an archive of puzzles with answers verified by their authors, structured tags, the asker's own spoiler-tagged hints, and timestamps.

We turn this archive into \bench{}. Three properties distinguish it from existing reasoning benchmarks. (1) \textbf{Scale with verified ground truth}: 23,777 puzzles, each with an author-accepted answer and a short reference answer taken from it; on the evaluation sets an LLM extracts the reference, checked on a 50-puzzle human review. (2) \textbf{Hints and tags}: 20\% of puzzles carry one or more hints written by the asker, allowing a with/without-hint ablation, and all puzzles carry community tags (201 distinct) for category analysis. (3) \textbf{A time axis}: creation dates let us report accuracy by posting period and hold out the 254 puzzles posted after the training cutoff of six of the nine models we evaluate (Section~\ref{sec:cutoff}), a slice that refreshes with every new data dump.

Our evaluation yields four findings. First, exact string match catches only about half of the correct answers (45--60\%, depending on the model); the remainder is decided by an LLM judge, which we validate with three judges from two families and find interchangeable (Section~\ref{sec:judge}). Second, on a shared 255-puzzle set GPT-6.1 reaches 74.8\%, ahead of Claude Opus 5.5 (66.1\%), and a hard core of 17 puzzles is solved by none of the nine models (Section~\ref{sec:overlap}). Third, model failures are strongly correlated, yet the union of nine models solves 91\% of the set, with most of the complementary coverage coming from a \emph{previous-generation} model rather than the second-best current one (Section~\ref{sec:overlap}). Fourth, a 100-puzzle multimodal pilot with an image-withheld control shows that the images in a third of the benchmark carry 24--35 points of accuracy (Section~\ref{sec:images}). Along the way we document two methodological points that generalize beyond this benchmark: timed-out rows are not missing at random and inflate the accuracy of slow models (Section~\ref{sec:errbias}), and the choice of reference answer matters as much as the choice of judge (Section~\ref{sec:refeffect}).

\section{Related Work}
\label{sec:related}

\paragraph{Reasoning benchmarks.} Broad suites such as BIG-bench \citep{srivastava2023beyond} and focused sets for arithmetic word problems \citep[GSM8K;][]{cobbe2021training}, competition mathematics \citep[MATH;][]{hendrycks2021measuring} and abstraction \citep[ARC;][]{chollet2019measure} have long tracked model reasoning, and newer benchmarks such as Humanity's Last Exam \citep{phan2026benchmark} rely on questions written by subject-matter experts. \bench{} instead draws on problems that hobbyists wrote for one another, each with an answer confirmed by its author.

\paragraph{Puzzles and riddles.} Existing puzzle benchmarks mostly isolate one genre or format: multiple-choice riddles and brain teasers \citep{lin-etal-2021-riddlesense,jiang-etal-2023-brainteaser}, interactive lateral-thinking puzzles \citep{huang-etal-2024-lateval}, generated logic grids \citep{lin2025zebralogic}, cryptic crosswords \citep{rozner2021decrypting,efrat-etal-2021-cryptonite,sadallah2024llms,saha-etal-2025-language}, the \textit{New York Times} Connections game \citep{todd2024missed,samdarshi-etal-2024-connecting} and the NPR Sunday Puzzle \citep{zhao2023solving,wu-etal-2025-reasoningweekly}. Multimodal puzzle sets cover abstract visual patterns \citep{chia-etal-2024-puzzlevqa}, rebuses \citep{gritsevskiy2024rebus}, jigsaw puzzles \citep{li2026jigshape} and problems from puzzle competitions \citep{wang2025enigmaeval}; \citet{giadikiaroglou-etal-2024-puzzle} survey the area. \bench{} spans these genres in one collection (201 community tags, free-form answers, images in a third of the puzzles) and attaches the asker's hints and a posting date to every puzzle.

\paragraph{Contamination and time-split evaluation.} Benchmark items leak into training data \citep{sainz-etal-2023-nlp,golchin2024time,deng-etal-2024-unveiling,deng-etal-2024-investigating}, and model accuracy has been shown to depend on whether a problem or dataset was released before the training cutoff \citep{roberts2023data,li2024task}. Live benchmarks respond by continually refreshing their questions \citep{white2025livebench,jain2025livecodebench,vu-etal-2024-freshllms}. \bench{} is static but dated: accuracy can be reported by posting period, and the source site added more than a thousand accepted-answer puzzles in each year from 2021 to 2025, so each quarterly dump yields a fresh slice.

\paragraph{LLM judges.} Using a strong LLM to grade free-form answers follows \citet{zheng2023judging}. Lexical metrics correlate poorly with human judgments of answer correctness \citep{lee-etal-2021-kpqa} and underestimate question-answering accuracy \citep{kamalloo-etal-2023-evaluating}, but LLM judges have their own failure modes: they are unreliable on hard, objectively gradable items \citep{tan2025judgebench} and can prefer their own outputs \citep{panickssery2024llm,wataoka2024self} or LLM-written summaries over human-written ones \citep{fang2026blind}. We therefore restrict the judge to an answer-equivalence decision, report three judges from two model families side by side, and measure own-family preference directly (Section~\ref{sec:judge}).

\paragraph{Stack Exchange as a data source.} Stack Exchange content has supplied data for community question answering \citep{hoogeveen2015cqadupstack} and, from Stack Overflow, for code generation and coding assistance \citep{yin2018learning,lai2023ds,shah2024stackeval}. In those settings an accepted answer is help for an asker who lacked it; on Puzzling SE the asker already knows the answer, which is what makes acceptance a verification signal.

\section{The \bench{} Dataset}
\label{sec:data}

\subsection{Source and collection}
We use the public Stack Exchange data dump of Puzzling Stack Exchange dated 2026-06-30, which contains 30,526 questions, 69,546 answers and 264,562 comments posted by 97,909 users between May 2014 and June 2026. Posts are CC BY-SA licensed; we retain post identifiers and author attribution fields for every record (Section~\ref{sec:ethics}). Images are referenced by URL in the dump; we fetched 42,566 of the 42,642 distinct image URLs in the dump's questions and answers (99.8\%). The 23,777 kept puzzles reference 28,868 of these URLs in their questions and accepted answers, and 28,825 of them were fetched and are released. Stack Exchange-hosted images posted before \mbox{2023-12-08} are also archived in the Stack Exchange Image Dataset \citep{StackExchangeImageDataset}.

\subsection{Filtering}
\label{sec:filter}
We keep a question if, and only if, (i) it was not closed by moderation, (ii) it has an answer the asker accepted, and (iii) that answer is present in the dump. Table~\ref{tab:funnel} shows the funnel. No tag-based ``is this a puzzle'' filter is applied; the site's own moderation is the only content filter. Because askers on this site almost always know the answer when they post, an accepted answer is a strong ground-truth signal; an author familiar with the site estimates it to be correct in roughly 99\% of cases, and that roughly 5\% of puzzles admit more than one valid answer (the asker can accept only one, but typically confirms other valid answers in the comments, which the dump also contains). We leave multiple-answer handling to future work.

\begin{table}[t]
\centering\small
\setlength{\tabcolsep}{4pt}
\begin{tabular}{@{}lrr@{}}
\toprule
Stage & Dropped & Remaining \\
\midrule
Questions in dump & & 30,526 \\
Closed by moderation & 2,694 (8.8\%) & 27,832 \\
No accepted answer & 4,054 (13.3\%) & 23,778 \\
Accepted answer missing & 1 & 23,777 \\
\midrule
\bench{} (v1) & & \textbf{23,777} (77.9\%) \\
\bottomrule
\end{tabular}
\caption{Filtering funnel from the raw dump to \bench{}.}
\label{tab:funnel}
\end{table}

For each puzzle we extract: the title; the question body (HTML converted to plain text, with each image replaced by an [image: alt-text] placeholder); the asker's hints, which on Puzzling SE are conventionally placed in spoiler blocks inside the question (4,825 puzzles, 20.3\%, carry at least one; 10,495 hint blocks in total, 2.2 per hinted puzzle); the full accepted answer; and a short \emph{reference answer} (Section~\ref{sec:reference}). We also record all community tags, the creation date, and whether the question or the accepted answer contains an image.

\subsection{Reference answers}
\label{sec:reference}
Scoring requires a short reference answer. Puzzling SE convention is that answerers state the solution in a spoiler block, so our first extractor took the first spoiler block of the accepted answer (available for 91\% of puzzles; the first paragraph otherwise). A manual audit showed this to be unreliable: on multi-spoiler answers, which make up 90\% of our evaluation set, the first block is a reasoning step rather than the solution about 10\% of the time (e.g.\ ``B and A in a Bacon cipher,'' where the solution, TRON, appears three blocks later). We therefore extract the reference with an LLM: Claude Opus 4.8 reads the question and the full accepted answer and returns the final answer the question asks for, its type (single, multi-part, construction, strategy), and the list of parts for multi-part answers, under an instruction to copy the answer faithfully and to exclude all justification. The extractor never sees any model output. On the 255-puzzle evaluation set it failed on 3 puzzles, which keep the spoiler-based reference. A human reviewed 50 randomly sampled extracted answers against the accepted answer: 46 were correct, 2 retained explanatory text as an additional answer part (corrected manually and logged), and 2 could not be verified; on the same 50 puzzles the spoiler-based reference was wrong 5 times. Section~\ref{sec:refeffect} reports the effect of the reference on scores. The released \texttt{short\_answer} field is the spoiler-based reference for all 23,777 puzzles; the LLM-extracted references cover the two evaluation sets (\pilot{} and \imgpilot{}, 355 puzzles) and are released with the per-row results.

\subsection{Statistics}
Table~\ref{tab:stats} summarizes the dataset. Two thirds of puzzles are text-only; 14,375 (60.5\%) have no image in either the question or the accepted answer, and this \emph{strict text-only} subset is the one our full-set experiments use. 7,907 puzzles (33.3\%) have an image in the question.

\begin{table}[t]
\centering\small
\begin{tabular}{lrr}
\toprule
 & $n$ & \% \\
\midrule
Puzzles & 23,777 & 100.0 \\
\quad text-only question & 15,870 & 66.7 \\
\quad strict text-only (Q and A) & 14,375 & 60.5 \\
\quad image in question & 7,907 & 33.3 \\
\quad image in accepted answer & 5,694 & 23.9 \\
With asker's hint(s) & 4,825 & 20.3 \\
Self-answered & 298 & 1.3 \\
\midrule
Posted $\le$2020 & 17,306 & 72.8 \\
Posted 2021--2024 & 4,863 & 20.5 \\
Posted 2025 & 1,027 & 4.3 \\
Posted 2026, before cutoff & 327 & 1.4 \\
Posted after cutoff (2026-05-01) & 254 & 1.1 \\
\midrule
Distinct tags & 201 & \\
Puzzles with $>$1 tag & 18,185 & 76.5 \\
Mean tags per puzzle & 2.34 & \\
\bottomrule
\end{tabular}
\caption{\bench{} statistics.}
\label{tab:stats}
\end{table}

\subsection{Categories}
\label{sec:tags}
Puzzling SE uses free-form community tags; \bench{} inherits all 201 of them. The most frequent are \textit{riddle} (22.6\%), \textit{mathematics} (17.5\%), \textit{knowledge} and \textit{word} (12.7\% each), \textit{logical-deduction} (11.4\%), \textit{enigmatic-puzzle} (10.9\%), \textit{visual} (8.9\%), \textit{wordplay} (8.7\%), \textit{pattern} (6.7\%) and \textit{cipher} (6.4\%). Tags are multi-label: 76.5\% of puzzles carry two or more, so per-tag accuracies (Appendix~\ref{app:tags}) count a puzzle under each of its tags. Any user with sufficient reputation can edit tags, so they are community-reviewed; an author familiar with the site puts their accuracy at roughly 80\%.

\subsection{Posting dates and the post-cutoff slice}
\label{sec:cutoff}
Every puzzle carries its creation timestamp, which lets us report accuracy by posting period (Section~\ref{sec:bydate}). We designate 2026-05-01 as the \emph{cutoff}. Six of the nine models we evaluate have a vendor-stated training cutoff of 2026-04-30 or earlier (GPT-6.1 Sol, GPT-5.6 Sol, Opus 4.8, Sonnet 5, Haiku 4.5 and Kimi K3); for Opus 5.5, Sonnet 5.5 and Grok 4.7 the vendors state June 2026, so the slice overlaps their cutoff. The 254 puzzles posted on or after 2026-05-01 (139 in May, 115 in June 2026) therefore postdate the training data of six of the nine models; 85 of them are strict text-only and 37 carry hints. Because the dump is a quarterly snapshot, this slice refreshes with each new dump. We regard it as a useful check rather than a decisive test: it is small (56 puzzles in our evaluation set), it is not post-cutoff for three of the models, and it cannot distinguish training exposure from a shift in puzzle difficulty over time.

\section{Evaluation Protocol}
\label{sec:protocol}

\subsection{Task format}
\label{sec:task}
A model receives a system instruction and a single user message containing the title and the question body, and must end its reply with a line of the form \texttt{ANSWER: <answer>}. Prompts are given verbatim in Appendix~\ref{app:prompts}. We extract the last \texttt{ANSWER:} line (tolerating ``Final answer:'' and Markdown decoration). If the reply has no such line, a second call shows the model the tail of its own output and demands a single committed answer within 100 tokens (requested, not enforced); if that also fails, the last non-empty line is taken and the row is flagged. The solve call requests a 4,096-token completion budget, but the shim through which all nine models are run (Section~\ref{sec:harness}) records \texttt{max\_tokens} without being able to enforce it, so the budget did not apply: 7--63\% of each model's \pilot{} replies are longer than 4,096 tokens, and none was truncated. The recovery path is recorded per row.

\subsection{Hints}
In the \emph{with-hints} condition the question is shown exactly as posted, including the asker's spoiler-tagged hints; in the \emph{without-hints} condition the hint blocks are removed. All results in Section~\ref{sec:results} use the with-hints condition.
\paragraph{Hint ablation.} Forty \pilot{} puzzles carry at least one hint block. To measure what the hints contribute, we re-ran GPT-6.1 Sol and Claude Opus 5.5 on all 255 puzzles with the hint blocks removed, under the same harness, effort, judge and reference (Table~\ref{tab:hints}). On the hinted puzzles, withholding the hints costs GPT-6.1 Sol 10.5 points and Opus 5.5 15.0, and the loss is concentrated on puzzles with two or more hint blocks ($-$23.5 and $-$26.3 points); puzzles with a single hint move by at most 4.8 points. The 215 unhinted puzzles, whose prompts are identical in both runs, give the run-to-run noise floor: their accuracy moves by at most 1.4 points, although 9--13 puzzles flip in each direction, so per-puzzle verdicts are noisier than the aggregates.

\begin{table}[t]
\centering\small\setlength{\tabcolsep}{5pt}
\begin{tabular}{@{}lrrrr@{}}
\toprule
 & $n$ & Hints & No hints & $\Delta$ \\
\midrule
\multicolumn{5}{@{}l}{\textit{GPT-6.1 Sol}} \\
\quad hinted puzzles & 38 & 55.3 & 44.7 & $-$10.5 \\
\quad\quad $\geq$2 hint blocks & 17 & 64.7 & 41.2 & $-$23.5 \\
\quad unhinted (noise floor) & 209 & 78.5 & 79.9 & $+$1.4 \\
\multicolumn{5}{@{}l}{\textit{Claude Opus 5.5}} \\
\quad hinted puzzles & 40 & 50.0 & 35.0 & $-$15.0 \\
\quad\quad $\geq$2 hint blocks & 19 & 47.4 & 21.1 & $-$26.3 \\
\quad unhinted (noise floor) & 214 & 69.2 & 68.7 & $-$0.5 \\
\bottomrule
\end{tabular}
\caption{Hint ablation on \pilot{}: accuracy (\%) with the asker's hints shown and withheld, paired on puzzles non-errored in both runs (Opus 4.8 judge, locked reference). Unhinted puzzles have identical prompts in both runs and measure run-to-run noise.}
\label{tab:hints}
\end{table}

\subsection{Two-tier scoring}
\label{sec:scoring}
\textbf{Tier 1} is normalized exact match against the reference answer: lowercase; HTML-unescape; strip lead-ins (``the answer is''), wrapping quotes and Markdown; map number words and thousands separators to digits (\textit{forty-two} $\to$ 42; 3,000 $\to$ 3000); hyphens to spaces; drop punctuation; and accept numeric equality under rational parsing (\textit{1/2} $=$ 0.5). \textbf{Tier 2} is invoked only when Tier 1 fails: an LLM judge sees the puzzle, the reference answer, the full accepted answer (as context) and the candidate, and returns \texttt{MATCH} or \texttt{NO\_MATCH} with a one-sentence reason (prompt in Appendix~\ref{app:prompts}). The judge is instructed not to solve the puzzle and not to grade reasoning; hedging between several answers or answering a different question is \texttt{NO\_MATCH}, and a multi-part puzzle matches only if every part the question asks for is answered. Judge calls use temperature 0 and a 200-token budget (requested, not enforced). \emph{Accuracy} is the fraction of non-errored rows accepted by either tier. Tier 1 alone accepts only 45--60\% of the answers finally scored correct (tier-1-only accuracy is 16.9--42.8\%, against 28.3--74.8\% with the judge), so the judge decides roughly half of every headline number; Section~\ref{sec:judge} validates it.

\subsection{Models and harness}
\label{sec:harness}
We evaluate nine models: GPT-6.1 Sol and GPT-5.6 Sol \citep{openai2026gpt61sol,openai2026gpt56}, Claude Opus 5.5, Opus 4.8, Sonnet 5.5, Sonnet 5 and Haiku 4.5 \citep{anthropic2025haiku45,anthropic2026opus48,anthropic2026opus55,anthropic2026sonnet5,anthropic2026sonnet55}, Kimi K3 \citep[Moonshot;][]{kimi2026k3} and Grok 4.7 \citep{xai2026grok47}. Open models (Llama-3.1-8B-Instruct, Qwen3-8B, Qwen3-14B; \citealp{grattafiori2024llama,yang2025qwen3}) serve as weak-judge baselines. Models are accessed through a local OpenAI-compatible shim in front of one of two agentic CLI harnesses: harness A for the 2026-10 runs and harness B for the earlier Claude runs. Harness A prepends a roughly 7.5K-character system prompt and a timestamp and fixes temperature at 0; harness B prepends its own system prompt. These settings are not under our control and apply identically to every model run through the same harness. Tool use is disabled; we verified that no tool schema reaches the model and record \texttt{tool\_requests} per row (0 on every row reported). Models sometimes still emit tool-call \emph{syntax} as plain text (7 Opus 5.5 rows, 8 Sonnet 5.5 rows); this is role-play, not execution, and does not affect answer extraction. Reasoning effort is \emph{medium} for every model that exposes it; Kimi K3 exposes no effort control and runs at its default. The earlier harness B runs (Opus 4.8, Sonnet 5, Haiku 4.5) had unbounded thinking and no effort setting. Every non-errored harness A row records the requested and served model identifier, effort, harness label, token usage, latency, cost and the judge identity; harness B rows, logged before this metadata existed, carry the harness label but no served model, effort or cost. Appendix~\ref{app:harness} lists all fields.

\subsection{Evaluation sets}
\label{sec:sets}
\textbf{\pilot{}} is a frozen set of 255 strict text-only puzzles with a spoiler-sourced short answer of at most 100 characters and a question of at most 4,000 characters, 40 of which carry at least one hint block: 200 drawn at random (seed 0) from the 9,508 eligible puzzles, plus all 56 eligible post-cutoff puzzles, one of which the random draw already contained ($200 + 56 - 1 = 255$). Its period mix is 163 / 28 / 5 / 3 / 56 for $\le$2020 / 2021--24 / 2025 / 2026-pre / post-cutoff. Every model is run on exactly these 255 puzzles via a frozen identifier list. \textbf{Full text-only} is the 14,375-puzzle strict text-only set, on which we run inexpensive models (Section~\ref{sec:full}). \textbf{\imgpilot{}} is a frozen 100-puzzle set of question-image puzzles with a text reference answer (Section~\ref{sec:images}).

\section{Results}
\label{sec:results}

\subsection{Main results on \pilot{}}
\begin{table*}[t]
\centering\small
\begin{tabular}{llrcrrrr}
\toprule
 & & \multicolumn{4}{c}{Accuracy (\%) by judge} & & \\
\cmidrule(lr){3-6}
Model & Harness / effort & Opus 4.8 & 95\% CI & Opus 5.5 & GPT-5.6 & Spread & Errored rows \\
\midrule
GPT-6.1 Sol       & A / medium    & \textbf{74.8} & 69.2--79.6 & \textbf{76.0} & \textbf{75.2} & 1.2 & 5 \\
Claude Opus 5.5   & A / medium    & 66.1 & 60.6--71.7 & 66.9 & 66.9 & 0.8 & 1 \\
Claude Opus 4.8   & B / unbounded & 62.9 & 56.9--69.0 & 62.5 & 62.9 & 0.4 & 23 \\
Kimi K3           & A / default   & 60.7 & 54.1--67.2 & 61.1 & 59.8 & 1.3 & 26 \\
GPT-5.6 Sol       & A / medium    & 60.0 & 53.7--65.9 & 60.0 & 60.4 & 0.4 & 0 \\
Grok 4.7          & A / medium    & 51.1 & 44.7--57.4 & 51.5 & 51.1 & 0.4 & 20 \\
Claude Sonnet 5   & B / unbounded & 49.8 & 43.9--56.1 & 49.4 & 49.0 & 0.8 & 16 \\
Claude Sonnet 5.5 & A / medium    & 47.8 & 42.0--53.7 & 46.3 & 47.1 & 1.6 & 0 \\
Claude Haiku 4.5  & B / unbounded & 28.3 & 22.8--33.5 & 29.1 & 28.0 & 1.2 & 1 \\
\bottomrule
\end{tabular}
\caption{\pilot{} results (255 puzzles, with hints, locked reference answers). Harness A: the 2026-10 runs at the stated reasoning effort; B: the earlier Claude runs with unbounded thinking (Section~\ref{sec:harness}). Accuracy over non-errored rows under each of three judges; the Opus 4.8 column is the paper's primary metric, and 95\% CI is its percentile bootstrap interval over 1,000 resamples of the non-errored puzzles. Spread: max minus min across judges. Errored rows: API errors or timeouts after retries (Section~\ref{sec:errbias}).}
\label{tab:main}
\end{table*}

Table~\ref{tab:main} gives the main result. GPT-6.1 Sol leads at 74.8\%, nearly 9 points ahead of Claude Opus 5.5 (66.1\%), which in turn is 3 points ahead of its predecessor Opus 4.8 (62.9\%). Kimi K3 (60.7\%) and GPT-5.6 Sol (60.0\%) form a second tier; Grok 4.7, Sonnet 5 and Sonnet 5.5 cluster at 48--51\%; Haiku 4.5 trails at 28.3\%. The ordering is identical under the two Claude judges and differs under the GPT-5.6 judge only by a swap of Kimi K3 and GPT-5.6 Sol, which are 0.6 points apart. Within the Claude family, the generation step from Sonnet 5 to Sonnet 5.5 brings no measurable gain (49.8 vs.\ 47.8, within the $\pm 3$-point standard error at $n{=}255$), while Opus 4.8 to Opus 5.5 brings +3.2.

\begin{figure}[t]
\centering
\includegraphics[width=\columnwidth]{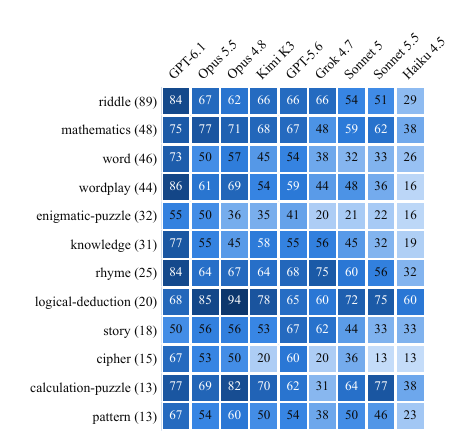}
\caption{Per-tag accuracy (\%) on \pilot{} for the twelve most frequent tags (Opus 4.8 judge), models in the order of Table~\ref{tab:main}; $n$ per tag in parentheses. Darker is higher. Table~\ref{tab:tags} gives the exact values.}
\label{fig:tags}
\end{figure}

Per-tag accuracy (Figure~\ref{fig:tags}; Appendix~\ref{app:tags}, Table~\ref{tab:tags}) shows a partly shared profile. \textit{Enigmatic-puzzle} is one of the two hardest of the twelve most frequent tags for every model, and \textit{cipher} is among the three hardest for all but GPT-5.6 Sol. The easy end differs by family: \textit{logical-deduction} is the easiest or second-easiest tag for the five Claude models and Kimi K3 (60--94\%), whereas GPT-6.1 Sol is strongest on \textit{wordplay}, \textit{riddle} and \textit{rhyme} (84--86\%). With 13--89 puzzles per tag, these figures are descriptive.

\subsection{Accuracy by posting date}
\label{sec:bydate}
\begin{figure}[t]
\centering
\includegraphics[width=\columnwidth]{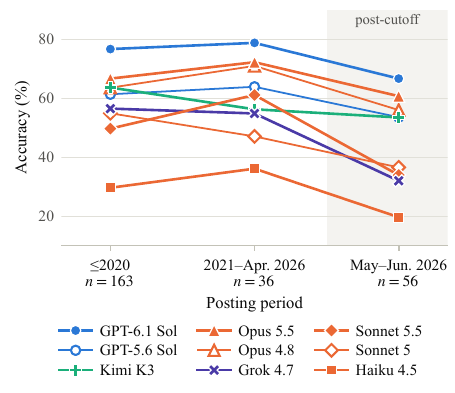}
\caption{Accuracy by posting period on \pilot{} (Opus 4.8 judge). The 2021--2024, 2025 and 2026 pre-cutoff buckets ($n{=}28$, 5, 3) are merged; $n$ counts puzzles, and each model's errored rows are excluded. Colour marks the developer and marker shape the model line; hollow markers and thinner lines mark the previous generation. One standard error is about 4, 8 and 7 points in the three periods.}
\label{fig:bydate}
\end{figure}

Figure~\ref{fig:bydate} plots accuracy against posting period for each model; because the 2025 and 2026 pre-cutoff buckets hold only five and three puzzles, we merge them with 2021--2024 into one 36-puzzle bucket. Two patterns are visible. First, accuracy does not decline steadily with recency before the cutoff: six of the nine models score higher on puzzles from 2021--April 2026 than on puzzles from 2020 or earlier. Second, every model scores lower on puzzles posted after the cutoff than on puzzles posted in 2020 or earlier, but the size of the drop varies widely: 6 points for Opus 5.5 (66.7 $\to$ 60.7), 8 for Opus 4.8 (63.6 $\to$ 56.0) and GPT-5.6 Sol (61.3 $\to$ 53.6), 10 for GPT-6.1 Sol (76.7 $\to$ 66.7), Kimi K3 (63.6 $\to$ 53.5) and Haiku 4.5 (29.6 $\to$ 19.6), against 16 for Sonnet 5.5 (49.7 $\to$ 33.9), 18 for Sonnet 5 (54.9 $\to$ 36.5) and 24 for Grok 4.7 (56.5 $\to$ 32.0). Two readings are compatible with this pattern: weaker models' scores on older puzzles may be inflated by training exposure, or recent puzzles may be harder in ways that affect weaker models more. For Opus 5.5, Sonnet 5.5 and Grok 4.7, whose stated cutoffs extend to June 2026 (Section~\ref{sec:cutoff}), the slice is not post-cutoff at all, so their drops cannot be read as exposure effects. With 56 post-cutoff puzzles (standard error about 7 points per model, about 8 for the difference from the $\le$2020 bucket) only the three largest drops exceed twice their standard error; we cannot separate the two readings, and we present the plot as a diagnostic that future dumps will sharpen rather than as a measurement of contamination. The earlier full-set run of Sonnet 5 (Section~\ref{sec:full}) shows the same direction for that model on a larger sample: 52.6\% for $\le$2020 ($n{=}749$), 43.1\% for 2021--24 ($n{=}130$) and 47.2\% for 2025 ($n{=}36$).

\subsection{Full-set results for inexpensive models}
\label{sec:full}
Running frontier models on all 14,375 text-only puzzles is costly; running inexpensive ones is not, and the full set is where the dataset's scale matters. Claude Haiku 4.5 completed 3,361 puzzles (23.4\% of the set) at 25.5\% accuracy (3,354 non-errored rows), Claude Sonnet 5 927 puzzles at 50.8\% and Opus 4.8 197 puzzles at 54.5\%. Full-set figures use the spoiler-based reference; with the same reference the three models score 29.5\%, 50.6\% and 65.5\% on \pilot{} (28.3\%, 49.8\% and 62.9\% with the locked reference of Table~\ref{tab:main}). Sonnet 5 thus agrees within a point and Haiku 4.5 within 4 points, which supports \pilot{} as a representative sample for these two models. Opus 4.8 scores 11 points lower on its 197 full-set puzzles than on \pilot{}, more than twice the standard error of the difference (about 5 points), so for this model the full-set run does not confirm the \pilot{} figure.

\subsection{Image puzzles (preliminary)}
\label{sec:images}
\imgpilot{} is a frozen set of 100 question-image puzzles (PNG or JPEG; 73 with a single image, one with 35) that have a text reference answer, stratified like \pilot{} (49 / 25 / 5 / 1 / 20 across the five periods; 19 with hints). Its category mix differs markedly from the text set: \textit{visual} (27), \textit{logical-deduction} (20), \textit{enigmatic-puzzle} (18), \textit{mathematics} (17), \textit{knowledge} (16), \textit{geometry} (11), \textit{grid-deduction} (10) and \textit{pattern} (9) dominate, and \textit{riddle} is absent. Images are passed to the model as attachments; the text prompt and the scorer are unchanged. As a control we run the same puzzles with the image withheld, so that the with-minus-without difference measures how much of each puzzle is carried by the image.

\begin{table}[t]
\centering\small
\begin{tabular}{@{}lrrrr@{}}
\toprule
 & & \multicolumn{2}{c}{Accuracy (\%)} & \\
\cmidrule(lr){3-4}
Model & $n$ & image & no image & $\Delta$ \\
\midrule
Claude Opus 5.5 & 99 / 98 & 54.5 & 31.6 & +23.5 \\
GPT-6.1 Sol & 93 / 96 & 64.5 & 28.1 & +34.8 \\
\bottomrule
\end{tabular}
\caption{\imgpilot{}: the same 100 image puzzles solved with the image attached and with it withheld (Opus 4.8 judge, effort medium, LLM-extracted reference). $n$ counts non-errored rows per condition; $\Delta$ is computed on puzzles non-errored in both. GPT-6.1 Sol timed out on 7 image-mode puzzles.}
\label{tab:images}
\end{table}

Table~\ref{tab:images} shows that the images carry a large part of the puzzle: withholding them costs Opus 5.5 24 points and GPT-6.1 Sol 35. Among the eight most frequent tags, the gain is largest on \textit{pattern} (+66.7 points for both models), \textit{knowledge} (+56.2 for Opus 5.5, +47.5 for GPT-6.1 Sol) and \textit{visual} (+44.4, +52.5). Opus 5.5 gains least on \textit{geometry} (0.0) and \textit{mathematics} (+17.6), which it often solves from the text alone (63.6\% and 58.8\% without the image), whereas GPT-6.1 Sol gains at least 34 points on every one of the eight. Both models remain well below their text-only \pilot{} accuracy, but the two sets differ in category mix, so this is not a like-for-like comparison. The 28--32\% accuracy \emph{without} the image shows that for roughly a third of these puzzles the text alone suffices; recall is not the explanation: for GPT-6.1 Sol, whose cutoff precedes the slice, no-image accuracy on the 20 post-cutoff \imgpilot{} puzzles is no lower than on the 49 posted in 2020 or earlier (31.6\% vs.\ 25.5\%), and Opus 5.5, whose cutoff overlaps the slice (Section~\ref{sec:cutoff}), shows the same pattern (36.8\% vs.\ 31.2\%). For 57 of the 100 puzzles the accepted answer also contains an image; we score against the text reference throughout, which may under-specify inherently visual answers (a completed grid, a drawn path). A vision-capable judge for such answers, validated against human ratings as \citet{liu2026humanaligned} do for image editing, is future work.

\section{Analysis}
\label{sec:analysis}

\subsection{How much does the judge matter?}
\label{sec:judge}
Because roughly half of each score comes from Tier 2, we re-judged every \pilot{} output with three judges from two families: Claude Opus 4.8 (the primary judge), Claude Opus 5.5 and GPT-5.6 Sol, all at default effort and temperature 0, and measure row-level agreement with Cohen's $\kappa$ \citep{cohen1960coefficient}. Table~\ref{tab:main} shows the three resulting accuracies side by side; no model's score moves by more than 1.6 points and the ranking is unchanged up to one adjacent swap of 0.6 points. Row-level agreement over the 1,540 judged rows is 96.4\% between Opus 4.8 and Opus 5.5 ($\kappa$ 0.87--0.96 across the nine solvers), 95.3\% between Opus 4.8 and GPT-5.6 Sol ($\kappa$ 0.82--0.93), and 96.9\% between Opus 5.5 and GPT-5.6 Sol ($\kappa$ 0.90--0.96). Agreement is lowest on GPT-6.1 Sol's answers (90.9\% between Opus 4.8 and GPT-5.6 Sol), whose longer replies give the judges more to disagree about, but $\kappa$ stays above 0.81 everywhere. We conclude that judge quality has plateaued at Opus 4.8 and keep it as the primary judge.

\paragraph{Own-family preference.} A judge might favour answers from its own model family \citep{panickssery2024llm}. We measure a judge's \emph{own-family bonus} as its accuracy on same-family solvers minus the mean of the other two judges on those solvers, and the analogous quantity on other families. All six values are below one point: Opus 4.8 gives Anthropic solvers +0.3 and other solvers $-$0.4; Opus 5.5 gives $-$0.1 and +0.8; GPT-5.6 Sol gives OpenAI solvers +0.2 and others $-$0.4. We find no own-family preference. Notably, under the earlier spoiler-based reference Opus 4.8's own-family bonus was +1.4: fragmentary references leave the judge latitude, and family preference appears in that latitude. A clean reference removes it.

\paragraph{Weak judges are biased, strong judges are not.} On the open-model outputs we additionally judged with Llama-3.1-8B and Qwen3-14B. Llama-8B is 14 points more lenient than Opus 4.8 on every solver family alike, Qwen3-14B 8 points stricter on every family alike: weak judges shift the scale but, in our data, do not favour their own family either.

\subsection{How much does the reference answer matter?}
\label{sec:refeffect}
We scored every \pilot{} output under three references: the original spoiler-based reference, and two rounds of LLM extraction (the second with the stricter no-justification instruction, plus the two manual corrections). Rankings are identical under the first and the final reference, and every model's accuracy moves by at most 2.8 points, downward for the stronger models (GPT-6.1 Sol 76.0 $\to$ 74.8, Opus 5.5 68.9 $\to$ 66.1, Opus 4.8 65.5 $\to$ 62.9). The direction is instructive: the fragmentary references had let the judge accept partial answers to multi-part puzzles by leaning on the full accepted answer; a complete reference enforces the all-parts rule that the judge prompt states. The net effect was small here because Tier 2 sees the full accepted answer, which rescued most bad references. Benchmarks whose judges see only a short reference would be affected more.

\subsection{Do models fail on the same puzzles?}
\label{sec:overlap}
Because every model saw the identical 255 puzzles, we can ask whether they fail on the same ones. Restricting to the 193 puzzles no model errored on, the answer is largely yes: given that GPT-6.1 fails a puzzle, Opus 5.5 fails it 73\% of the time, against a 28\% base rate; pairwise $\phi$ coefficients of the failure indicators range from 0.26 to 0.67, highest between the two Sonnets.

\begin{table}[t]
\centering\small
\begin{tabular}{@{}lrrr@{}}
\toprule
Step & Model added & Adds & Union acc.\ (\%) \\
\midrule
1 & GPT-6.1 Sol & 148 & 76.7 \\
2 & Claude Opus 4.8 & +18 & 86.0 \\
3 & GPT-5.6 Sol & +6 & 89.1 \\
4 & Grok 4.7 & +3 & 90.7 \\
5 & Claude Opus 5.5 & +1 & 91.2 \\
6--9 & (the other four) & +0 & 91.2 \\
\bottomrule
\end{tabular}
\caption{Greedy ensemble on the 193-puzzle common subset of \pilot{}: at each step the model adding the most newly solved puzzles joins the union. Opus 5.5, the second-best single model, joins fifth and adds one puzzle; Kimi K3, Sonnet 5, Sonnet 5.5 and Haiku 4.5 add none.}
\label{tab:overlap}
\end{table}

Yet the models are not interchangeable. An oracle that picks the best model per puzzle solves 91.2\% of the common set; 25.9\% are solved by all nine. Coverage-oriented routing exploits this complementarity, choosing model subsets that maximize the chance that at least one model is correct \citep{wei2026flexrouter}. The complementary coverage comes from unexpected places (Table~\ref{tab:overlap}): after GPT-6.1, the model that adds most is the previous-generation Opus 4.8 (+18 puzzles), not the current Opus 5.5, which would add 12 at that step and adds only one once Opus 4.8, GPT-5.6 Sol and Grok 4.7 have joined; no puzzle is solved by Opus 5.5 alone. Overlap is high but not nested: within a family the newer model solves 85--90\% of what its predecessor solved (Opus 5.5 solves 112 of Opus 4.8's 127 puzzles, Sonnet 5.5 94 of Sonnet 5's 110, GPT-6.1 Sol 118 of GPT-5.6 Sol's 131), and across families GPT-6.1 Sol solves 127 of the 139 puzzles that Opus 5.5 solves.

\paragraph{The hard core.} Seventeen puzzles (8.8\% of the common set) are solved by no model. Their tags skew toward \textit{steganography} (3 of the 17), \textit{cryptic-clues} (3), \textit{story} (4), \textit{enigmatic-puzzle} (4), \textit{cipher} (2) and \textit{mathematics} (5, among them \textit{combinatorics}, \textit{number-theory} and \textit{information-theory} puzzles); only three carry the \textit{riddle} tag, out of 75 riddle-tagged puzzles in the common set. Plain riddles, the largest category, are the solved genre; puzzles that hide information in the presentation itself are the unsolved one. Appendix~\ref{app:hardcore} lists all seventeen.

\subsection{Errored rows are not missing at random}
\label{sec:errbias}
Our harness enforces a 600-second limit per call. Kimi K3 and Grok 4.7 each exceeded it on 20 \pilot{} puzzles (Kimi K3 lost six more rows to empty replies), and excluding such rows inflated their accuracy: when we recovered 12 Kimi rows and 26 Grok rows by re-running, Kimi dropped from 65.0\% to 62.9\% and Grok from 55.0\% to 50.6\% (spoiler-based reference). Models time out on the puzzles they find hardest, so ``accuracy over non-errored rows'' is biased upward for slow models in proportion to their error rate. We report errored-row counts in every table for this reason, and recommend that benchmarks of long-reasoning models do the same.

Appendix~\ref{app:format} reports format compliance and the relation between output length and accuracy (Figure~\ref{fig:tokens}).

\section{Conclusion}
\label{sec:conclusion}
We presented \bench{}, 23,777 puzzles from Puzzling Stack Exchange whose answers were verified by the people who wrote them, each with hints, tags, a posting date and a short reference answer, LLM-extracted for the evaluation sets and checked on a 50-puzzle human review. On a shared 255-puzzle set, nine frontier and open models span 28--75\% under a two-tier scorer whose LLM judge we validated with three judges from two families: the judge moves no model by more than 1.6 points and shows no own-family preference, whereas the choice of reference answer moved the stronger models by up to 2.8 points. Model failures are strongly correlated, a hard core of 17 puzzles resists all nine models, and the complementary coverage in an ensemble comes from a previous-generation model rather than the second-best current one. The post-cutoff slice shows larger drops for weaker models, a pattern we present as a diagnostic that each quarterly dump will sharpen, and withholding a puzzle's image costs the two best models 24--35 points. We release the data, the scoring pipeline and every per-row output so that these analyses can be rerun as models and dumps change.
\label{end:body}

\section{Limitations}
\textbf{Judge dependence.} Although three judges agree closely, all three are frontier models; a puzzle whose reference answer is wrong, or whose candidate answer is correct but phrased in a way all three misread, is scored wrong by all of them. \textbf{LLM-derived references.} Reference answers are extracted by an LLM from the human-accepted answer. The extractor never sees model outputs, and a human review found it correct on 46 of 50 samples with two over-specified answers corrected, but the remaining error rate is nonzero and three puzzles fall back to the spoiler-based reference. \textbf{Harness.} Models were run through agentic CLIs that inject their own system prompts, and harness A fixes temperature at 0; results may differ from direct API access, and the earlier harness B runs had unbounded reasoning while the later harness A runs used medium effort, so cross-harness comparisons (Opus 4.8 vs.\ 5.5; Sonnet 5 vs.\ 5.5) carry this confound. \textbf{Sample size.} \pilot{} is 255 puzzles (standard error about 3 points) and its post-cutoff slice 56 (about 7 points); differences below those margins are not interpretable. \textbf{Contamination.} The post-cutoff slice postdates the stated training cutoff of six of the nine models but not of Opus 5.5, Sonnet 5.5 and Grok 4.7 (Section~\ref{sec:cutoff}); it does not control for re-posted variants of older puzzles, and it cannot separate exposure from difficulty drift. \textbf{Ground truth.} Accepted answers are the asker's choice, not a proof; the community occasionally accepts an incorrect answer, and about 5\% of puzzles admit more than one valid answer that we do not yet credit. \textbf{Images.} Only puzzles with text reference answers are evaluated in the image pilot. \textbf{Tags.} Community tags are multi-label and uneven in granularity; per-tag figures are descriptive, not a taxonomy. \textbf{Generality.} Whether performance on \bench{} tracks general reasoning ability, rather than puzzle-specific skill, is not established here.

\section{Ethical considerations}
\label{sec:ethics}
Puzzling Stack Exchange content is licensed CC BY-SA: version 3.0 for content contributed before 2018-05-02 and version 4.0 from that date on \citep{stackexchange-licensing}. As recorded in the dump, 9,019 of the 23,777 questions (37.9\%) and 9,403 of their accepted answers (39.5\%) are under CC BY-SA 3.0, and the rest, including every post created or last edited since 2018-05-02, under 4.0; the dump as a whole is distributed under CC BY-SA 4.0. We redistribute question and answer text with post identifiers and author display names so that attribution is preserved, under the same licence. Images are redistributed as fetched, with their source URLs. No personal data beyond public display names is included. Evaluating on this benchmark incurs API cost; the nine-model \pilot{} evaluation with three judges, the image pilot and the reference extraction cost roughly \$420. The dataset text, hints, tags, dates and reference answers, together with every per-row model output and judge verdict, are released under CC BY-SA at \url{https://huggingface.co/datasets/puzzling-bench/puzzling-bench}, and the code at \url{https://github.com/a1bhinav/puzzling-bench}.

\bibliography{custom}

\appendix
\section{Prompts}
\label{app:prompts}

\begin{table*}[t]
\centering\small
\setlength{\tabcolsep}{3.7pt}
\begin{tabular}{@{}lrrrrrrrrrr@{}}
\toprule
Tag & $n$ & GPT-6.1 & Opus 5.5 & Opus 4.8 & Kimi K3 & GPT-5.6 & Grok 4.7 & Sonnet 5 & Sonnet 5.5 & Haiku 4.5 \\
\midrule
\textit{riddle} & 89 & \textbf{84.3} & 67.4 & 61.7 & 66.3 & 66.3 & 66.3 & 53.5 & 50.6 & 29.2 \\
\textit{mathematics} & 48 & 75.0 & \textbf{77.1} & 71.1 & 68.2 & 66.7 & 47.7 & 58.7 & 62.5 & 37.5 \\
\textit{word} & 46 & \textbf{73.3} & 50.0 & 57.1 & 45.2 & 54.3 & 38.1 & 31.8 & 32.6 & 26.1 \\
\textit{wordplay} & 44 & \textbf{85.7} & 61.4 & 68.6 & 54.1 & 59.1 & 43.9 & 47.6 & 36.4 & 15.9 \\
\textit{enigmatic-puzzle} & 32 & \textbf{55.2} & 50.0 & 35.7 & 35.0 & 40.6 & 20.0 & 21.4 & 21.9 & 15.6 \\
\textit{knowledge} & 31 & \textbf{76.7} & 54.8 & 44.8 & 58.3 & 54.8 & 56.0 & 44.8 & 32.3 & 19.4 \\
\textit{rhyme} & 25 & \textbf{84.0} & 64.0 & 66.7 & 64.0 & 68.0 & 75.0 & 60.0 & 56.0 & 32.0 \\
\textit{logical-deduction} & 20 & 68.4 & 85.0 & \textbf{94.1} & 77.8 & 65.0 & 60.0 & 72.2 & 75.0 & 60.0 \\
\textit{story} & 18 & 50.0 & 55.6 & 56.2 & 52.9 & \textbf{66.7} & 62.5 & 43.8 & 33.3 & 33.3 \\
\textit{cipher} & 15 & \textbf{66.7} & 53.3 & 50.0 & 20.0 & 60.0 & 20.0 & 36.4 & 13.3 & 13.3 \\
\textit{calculation-puzzle} & 13 & 76.9 & 69.2 & \textbf{81.8} & 70.0 & 61.5 & 30.8 & 63.6 & 76.9 & 38.5 \\
\textit{pattern} & 13 & \textbf{66.7} & 53.8 & 60.0 & 50.0 & 53.8 & 38.5 & 50.0 & 46.2 & 23.1 \\
\midrule
All puzzles & 255 & \textbf{74.8} & 66.1 & 62.9 & 60.7 & 60.0 & 51.1 & 49.8 & 47.8 & 28.3 \\
\bottomrule
\end{tabular}
\caption{Per-tag accuracy (\%) on \pilot{} for the twelve most frequent tags (Tier 1 plus the Opus 4.8 judge, locked reference answers), with the best model per tag in bold (GPT-6.1 and GPT-5.6 are the Sol models). $n$ is the number of puzzles carrying the tag; a puzzle counts under each of its tags. Accuracy is over each model's non-errored rows, so denominators are smaller than $n$ for models with errored rows.}
\label{tab:tags}
\end{table*}

\paragraph{Solver system prompt.}
\begin{quote}\small\ttfamily\raggedright
You are solving a puzzle from Puzzling Stack Exchange. Think it through, but keep the reasoning compact. Your reply MUST end with exactly one line of the form:\\
ANSWER: <your answer>\\
Put only the answer on that line, with no explanation. Never end without the ANSWER line.
\end{quote}

\paragraph{Solver user prompt.}
\begin{quote}\small\ttfamily\raggedright
Title: \{title\}\\[2pt]
\{question body, with or without hint blocks\}\\[2pt]
End your reply with one line: ANSWER: <your answer>
\end{quote}

\paragraph{Forced follow-up} (sent when the first reply has no \texttt{ANSWER:} line; 100-token budget, requested, not enforced).
\begin{quote}\small\ttfamily\raggedright
-{}-{}-\\
Your earlier attempt at this puzzle was cut off. The tail of your notes:\\[2pt]
\{last lines of the first reply\}\\[2pt]
-{}-{}-\\
Do NOT continue the notes and do NOT reason further. Commit to your single best final answer now, guessing if necessary, and reply with exactly one line:\\
ANSWER: <answer>
\end{quote}

\paragraph{Judge system prompt.}
\begin{quote}\small\ttfamily\raggedright
You grade answers to puzzles. You will be shown a puzzle, the reference answer written by a human (a short answer plus the full accepted answer for context), and a candidate answer produced by an AI.\\[2pt]
Decide only whether the candidate names the SAME solution as the reference, for what the puzzle asks. Do not solve the puzzle yourself and do not grade the reasoning. Different wording, formatting, units, or extra explanation are fine if the solution is the same. If the candidate gives several different final answers, hedges between options, or answers a different question, it is NO\_MATCH. If the puzzle has several parts, MATCH only if every part matches.\\[2pt]
Reply with exactly two lines:\\
VERDICT: MATCH or NO\_MATCH\\
REASON: one sentence
\end{quote}

\paragraph{Judge user prompt.}
\begin{quote}\small\ttfamily\raggedright
PUZZLE:\\ \{question, truncated to 6,000 chars\}\\[2pt]
REFERENCE SHORT ANSWER:\\ \{reference answer\}\\[2pt]
REFERENCE FULL ANSWER (context):\\ \{accepted answer, truncated to 4,000 chars\}\\[2pt]
CANDIDATE ANSWER:\\ \{extracted ANSWER line\}
\end{quote}

\paragraph{Reference extraction system prompt} (temperature 0; 400-token budget, requested, not enforced).
\begin{quote}\small\ttfamily\raggedright
You extract final answers from puzzle solutions. You will see a puzzle and the accepted human-written answer to it. The accepted answer usually contains the solution plus reasoning, hints, or worked steps.\\[2pt]
Return ONLY a JSON object with these fields:\\
\hspace*{1em}"canonical\_answer": the final answer as one concise line, exactly as a grader should compare against (a word, number, phrase, expression, or short list). No reasoning. If the puzzle has several parts, join the parts with " | ".\\
\hspace*{1em}"answer\_type": one of "single", "multi\_part", "strategy", "construction", "unclear".\\
\hspace*{3em}single = one definite answer; multi\_part = several sub-answers all required; strategy = a procedure or proof; construction = a grid, arrangement, or list too long for one line; unclear = no definite final answer stated.\\
\hspace*{1em}"parts": for multi\_part, a list of the sub-answers in order; otherwise [].\\
\hspace*{1em}"notes": one short sentence if anything is ambiguous, else "".\\[2pt]
The canonical answer must contain ONLY the final answer(s) the question explicitly asks for, copied as faithfully as possible from the accepted answer. No justification, no explanation, no restatement of the method. Use multi\_part ONLY when the question itself asks for several things; never treat an explanation as a part. Do not correct, expand, or complete words; copy them exactly.\\[2pt]
Do not solve the puzzle yourself; extract what the accepted answer says. No markdown, no code fences.
\end{quote}

\paragraph{Reference extraction user prompt.}
\begin{quote}\small\ttfamily\raggedright
PUZZLE TITLE: \{title\}\\[2pt]
PUZZLE:\\ \{question, truncated to 5,000 chars\}\\[2pt]
ACCEPTED ANSWER:\\ \{accepted answer, truncated to 5,000 chars\}
\end{quote}

\section{Harness and provenance}
\label{app:harness}
\textbf{Shim.} Requests use the OpenAI chat-completions schema and are translated to a non-interactive CLI invocation: the prompt is passed on standard input, and the model identifier, the reasoning effort (when supported) and any image attachments are passed as options; tool use is disabled. The CLI's JSONL event stream is parsed for the assistant message and token and cost usage. Auth and quota failures are returned as HTTP 402 and not retried. \textbf{Per-row fields.} \texttt{model}, \texttt{served\_model}, \texttt{reasoning\_effort}, \texttt{run\_meta} (harness label, effort served, \texttt{tool\_requests}, latency, input/output/reasoning/cached tokens, cost, attachments sent), \texttt{prompt\_tokens}, \texttt{completion\_tokens}, \texttt{max\_tokens}, \texttt{finish\_reason}, \texttt{calls}, \texttt{extracted\_answer}, \texttt{answer\_source}, \texttt{tier1}, \texttt{judge\_used}, \texttt{verdict}, \texttt{reason}, \texttt{judge\_model}, \texttt{judge\_served\_model}, \texttt{judge\_meta}, \texttt{judge\_usage}, \texttt{correct}, \texttt{error}. For harness B rows and errored rows, \texttt{run\_meta} holds only the harness label. \textbf{Known harness caveats.} Harness A injects an approximately 7.5K-character system prompt and a timestamp and sets temperature 0; harness B injects its own system prompt. Harness B did not report prompt-token usage, so prompt-token counts for the Opus 4.8, Sonnet 5 and Haiku 4.5 runs are unreliable (completion counts are not affected). Kimi K3 rejects any effort setting. Rows from the Haiku 4.5 full run that were re-generated after an authentication failure (266 rows) carry the harness A label and score 25.2\% against 25.5\% for the original rows.

\textbf{Cost.} One judge verdict costs about \$0.021; one \pilot{} puzzle including the judge about \$0.059 (model-dependent); one reference extraction about \$0.02 (estimated from the judge's recorded usage; extraction usage was not logged).

\section{Per-tag results}
\label{app:tags}
Table~\ref{tab:tags} gives accuracy on the twelve most frequent \pilot{} tags for each model.

\section{The hard core}
\label{app:hardcore}
The seventeen \pilot{} puzzles solved by none of the nine models (post identifier: tags):
8076 (mathematics, lateral-thinking, calculation-puzzle, situation);
12411 (mathematics, strategy, information-theory);
23586 (crossword-clues, seasonal);
24227 (pattern, wordplay, cryptic-clues);
28064 (mathematics, combinatorics);
32412 (enigmatic-puzzle, story, cipher);
42168 (cryptic-clues);
46930 (story, steganography, cryptic-clues);
59412 (enigmatic-puzzle, steganography, situation);
62112 (mathematics, number-theory);
63878 (riddle, word, rhyme);
67349 (riddle, mathematics);
94165 (word, story);
128441 (word, knowledge, enigmatic-puzzle);
131053 (word, enigmatic-puzzle);
137870 (cipher, story, steganography);
138092 (riddle, poetry, limerick).

\section{Format compliance and output length}
\label{app:format}
\begin{figure}[t]
\centering
\includegraphics[width=\columnwidth]{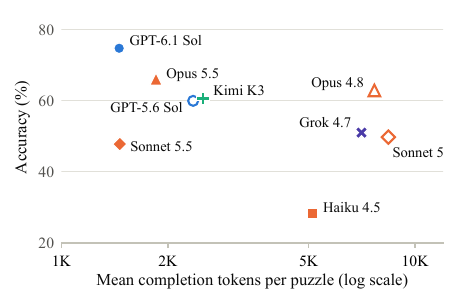}
\caption{\pilot{} accuracy (Opus 4.8 judge) against mean completion tokens per non-errored puzzle, solve call plus any forced follow-up. Colour and marker as in Figure~\ref{fig:bydate}; hollow markers are the previous generation; the three harness B runs (Opus 4.8, Sonnet 5, Haiku 4.5) had unbounded thinking.}
\label{fig:tokens}
\end{figure}

Grok 4.7 produced the required \texttt{ANSWER:} line on its first reply for only 94 of its 235 non-errored puzzles; the forced follow-up recovered an answer in all remaining 141 cases. Every other model complied on the first reply for 96--100\% of puzzles. Because the shim records the 4,096-token budget without enforcing it (Section~\ref{sec:task}), no reply on \pilot{} was truncated even though many ran far past it: mean completion length per non-errored puzzle ranges from roughly 1.5K tokens (GPT-6.1 Sol, Sonnet 5.5) to 7K (Grok 4.7) among the harness A models, and reaches about 8K in the unbounded-thinking harness B runs of Opus 4.8 and Sonnet 5. Figure~\ref{fig:tokens} plots accuracy against mean completion length: the two are not monotonically related, and the two models with the shortest replies, GPT-6.1 Sol and Sonnet 5.5, sit at the top and near the bottom of the harness A group.

\section{Example puzzles}
\label{app:examples}
Three \pilot{} puzzles, chosen for brevity, with the locked reference answer and the extracted \texttt{ANSWER} lines of the two best models (Opus 4.8 judge; Tier 1 is the normalized exact match of Section~\ref{sec:scoring}). None of the three carries a hint.

\paragraph{Solved by all nine models.} \textit{End of every race, Who am I?} (post 14532; \textit{riddle})
\begin{quote}\small
What is the beginning of eternity,\\
The end of time and space,\\
The beginning of every end,\\
And the end of every race?
\end{quote}
{\raggedright Question by \href{https://puzzling.stackexchange.com/users/4312}{coDe murDerer}, \href{https://puzzling.stackexchange.com/q/14532}{\nolinkurl{puzzling.stackexchange.com/q/14532}}, CC BY-SA 3.0.\par}
\noindent Reference: \emph{letter E}. GPT-6.1 Sol: ``E'' (judge: \texttt{MATCH}). Opus 5.5: ``The letter E'' (Tier 1 match).

\paragraph{Solved by GPT-6.1 Sol only.} \textit{An overlapping Riley Riddle} (post 63955; \textit{riddle, word, wordplay, rhyme})
\begin{quote}\small
My attempt at a Riley Riddle:\\
My prefix is a limit.\\
My suffix is quite unlit.\\
My infix shows what's done-{}-\\
but I'm afraid I don't have one.\\
What am I?
\end{quote}
{\raggedright Question by \href{https://puzzling.stackexchange.com/users/43978}{Carl Schildkraut}, \href{https://puzzling.stackexchange.com/q/63955}{\nolinkurl{puzzling.stackexchange.com/q/63955}}, CC BY-SA 4.0.\par}
\noindent Reference: \emph{Infinite}. GPT-6.1 Sol: ``Infinite'' (Tier 1 match). Opus 5.5: ``Unlimited'' (judge: \texttt{NO\_MATCH}).

\paragraph{Hard core (solved by no model).} \textit{I'm Gone Before You Know It. What Am I?} (post 67349; \textit{riddle, mathematics})
\begin{quote}\small
Here's a crazy riddle for you. Figure out what I am, if you can:\\
I'm gone before you know it, yet I go on and on and on\ldots\\
Can you tell me what I am?
\end{quote}
{\raggedright Question by \href{https://puzzling.stackexchange.com/users/4491}{bob}, \href{https://puzzling.stackexchange.com/q/67349}{\nolinkurl{puzzling.stackexchange.com/q/67349}}, CC BY-SA 4.0.\par}
\noindent Reference: \emph{Pi/pie}. GPT-6.1 Sol: ``Time'' (judge: \texttt{NO\_MATCH}). Opus 5.5: ``Time (the present moment)'' (judge: \texttt{NO\_MATCH}).

\end{document}